\documentclass[journal]{IEEEtran}
\usepackage{cite}
\usepackage{hyperref}
\usepackage{times}
\usepackage{epsfig}
\usepackage{color}
\usepackage{graphicx}
\usepackage[utf8]{inputenc}
\usepackage{amsmath}
\usepackage{amssymb}
\usepackage{tabularx}
\usepackage{authblk}
\newcommand{\tabincell}[2]{\begin{tabular}{@{}#1@{}}#2\end{tabular}}

\begin{document}

\title{Tone Mapping Based on Multi-scale Histogram Synthesis}


\author[2]{Jie Yang  \thanks{Corresponding author e-mail: yangjie@westlake.edu.cn}}
\author[1]{Ziyi Liu} 
\author[1]{Ulian Shahnovich}
\author[1]{Orly Yadid-Pecht}

\affil[1]{I2Sense lab, University of Calgary, Calgary T2N 1N4, Canada}
\affil[2]{Westlake University, Hangzhou  310024, China}

{}

\maketitle

\begin{abstract}
In this paper, we present a novel tone mapping algorithm that can be used for displaying wide dynamic range (WDR) images on low dynamic range (LDR) devices. The proposed algorithm is mainly motivated by the logarithmic response and local adaptation features of the human visual system (HVS). HVS perceives luminance differently when under different adaptation levels, and therefore our algorithm uses functions built upon different scales to tone map pixels to different values. Functions of large scales are used to maintain image brightness consistency and functions of small scales are used to preserve local detail and contrast. An efficient method using local variance has been proposed to fuse the values of different scales and to remove artifacts. The algorithm utilizes integral images and integral histograms to reduce computation complexity and processing time. Experimental results show that the proposed algorithm can generate high brightness, good contrast and appealing images that surpass the performance of many state-of-the-art tone mapping algorithms. This project is  available at https://github.com/jieyang1987/Tone-Mapping-Based-on-Multi-scale-Histogram-Synthesis.
\end{abstract}

\begin{IEEEkeywords}
Wide dynamic range image (WDR), tone mapping, local adaptation, multiscale, fusion.
\end{IEEEkeywords}

\section{Introduction}
The dynamic range is defined as the ratio of the intensity of the brightest point to the intensity of the darkest point in a scene or image. The dynamic range of a natural scene can go beyond 
120 dB, which exceeds the dynamic range of almost all modern image sensors. The traditional approach for capturing a wide dynamic scene is to take multiple images with different exposures and fuse all these images together to form a wide dynamic range (WDR) image \cite{debevec1997recovering}. However, progress in image sensors has made direct capturing of wide dynamic range images possible. {Logarithmic response sensors \cite{bouvier2014logarithmic,kavadias2000logarithmic}, multimode sensors \cite{bae2016linear,storm2006extended}, capacitance adjustment sensors \cite{decker1998256,fossum2005high} and other technologies such as \cite{kronander2014unified} offer a possibility to extend the sensor dynamic range.}
However, the dynamic range of traditional display devices such as LCD, CRT and LED are usually limited to 8 bits, and hence in many cases it is impossible to properly reproduce the WDR image on the display directly. In order to close the gap of dynamic range difference between WDR image and LDR display devices, tone mapping algorithms were developed. Tone mapping algorithms, also called tone mapping operators (TMO), serve two purposes: the first one is to compress the WDR image to the dynamic range of display devices, and the second one is to generate {high quality images for different application scenarios.}
Based on the application, tone mapped images may require different attributes. For example, in photography human preference is the first priority, while in security and machine vision satisfaction of certain image property or visibility are more important requirements.
 
A global tone mapping process is to apply a single global function to all pixels in the image where an identical pixels will be given {an} identical output value within the range of the display devices. Tumblin and Rushmeier \cite{tumblin1993tone} and Ward \cite{ward1994contrast} were the early researchers who developed  global operators for tone mapping. Drago et al. \cite{drago2003adaptive} proposed an adaptive logarithmic mapping method that can change the base of the logarithmic function based on the brightness. Recently, Hor{\'e} et al.\cite{hore2014statistical,hore2014new} proposed an hybrid tone mapping algorithm and its hardware implementation \cite{ambalathankandy2016fpga} that takes into account image local statistics. Such kind of algorithms can also be found in \cite{larson1997visibility,qiu2005optimal,tsai2012fast}. In general, global tone mapping algorithms are computationally easy to implement and mostly ``artifacts"-free, and they have unique advantages in hardware implementations.
However, the tone mapped images of these algorithms may suffer low brightness, low contrast or loss of details due to the global compression of the dynamic range. Local TMOs become the mainstream of tone mapping. Inspired by some features of the human visual system, some local tone mapping algorithms \cite{reinhard2005dynamic,van2006encoding,spitzer2003biological} try to mimic the dynamic range compression process of our photoreceptors. 
Some researchers solve the WDR compression as a constrained optimization problem. Mantiuk et al. \cite{mantiuk2008display} considered the tone mapping as a minimum visible distortion problem. Ma et al. \cite{ma2014high} proposed a tone mapping method by optimizing the tone mapped image quality index. However, optimizing a single metric can hardly guarantee the best results. Additionally, solving constrained optimization problem is computationally expensive and difficult to implement in real-time. 
In recent years, various algorithms emerged based on the Retinex theory \cite{land1971lightness, herscovitz2004modified,meylan2006high}. Edge preserving filters  were used to separate the WDR image into illuminance and reflectance channels. The illuminance channel was regarded as base layer whose information was believed less important for our visual system, and hence its dynamic range was greatly compressed. On the other hand, reflectance channel was treated as detail layer whose information was mostly preserved during tone mapping. 
The tone mapped images using these edge-preserving filters give state-of-the-art quality \cite{durand2002fast,farbman2008edge,he2010guided, gu2013local, paris2015local}.

\begin{figure}[tbh]
\begin{center}
   \includegraphics[scale=0.35]{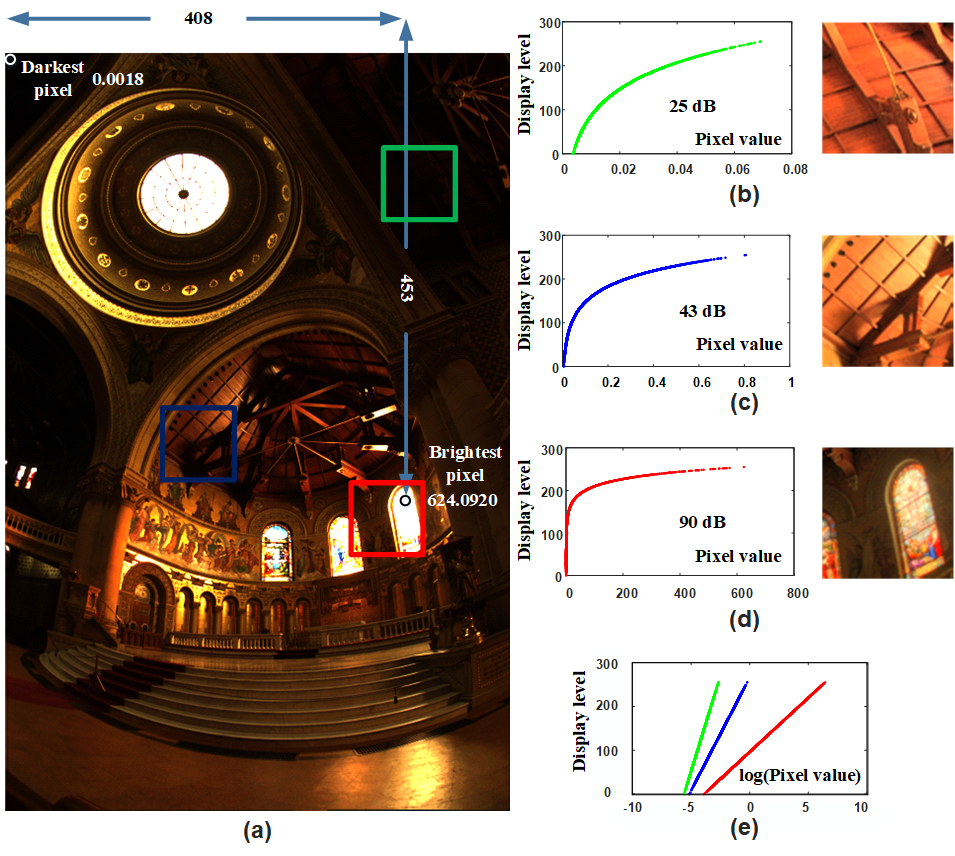}
\end{center}
   \caption{Tone mapping local blocks from Memorial Church image. {(a) Original WDR image (tone mapped for display). 
(b) Logarithmic tone curve that maps the green block in (a) and the result block image.
(c) Logarithmic tone curve that maps the blue block in (a) and the result block image. 
(d) Logarithmic tone curve that maps the red block in (a) and the result block image. 
(e) Tone mapping curves of green, blue and red blocks in logarithmic domain. 
Memorial radiance map courtesy of Paul Debevec, University of California at Berkeley.}}
\label{Local_Adaptation}
\end{figure}

\begin{figure}[tbh]
\begin{center}
   \includegraphics[scale = 0.65]{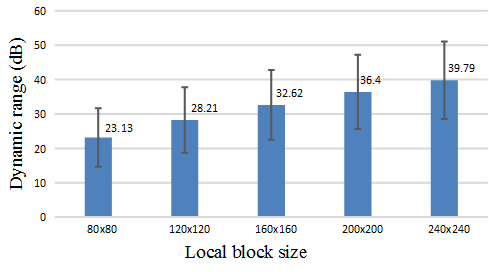}
\end{center}
   \caption{Statistic measurements of the local image block dynamic range.}
\label{Fig_2}
\end{figure}

{We have briefly introduced our new tone mapping algorithm based on multi-scale histogram synthesis (MS-Hist) in \cite{yang2017multi}. In this paper, we expand the work presented to include motivation, detailed description and analysis of the algorithm and its optimization process. Moreover, comprehensive experiments and analysis including a developed iOS application evaluation, objective and subjective assessments were carried out and presented here.}
{The MS-Hist algorithm can generate tone curves for every pixel of a WDR image based on the local histograms. Some excellent works have been done using local histogram based tone curves. For example, Duan \textit{et al.} \cite{duan2010tone} have proposed a tone mapping algorithm based on histogram adjustment which combines linear mapping and histogram equalized mapping. The adjustment method is applied to non-overlapping local regions and the final value of a pixel is a weighted average of the results from tone mapping functions of adjacent regions. Boschetti \textit{et al.} \cite{boschetti2010high} reported an algorithm derived from the Contrast Limited Adaptive Histogram Equalization (CLAHE) technique, the algorithm can adaptively control the local contrast limit, the adaptation is achieved by combining the local mean and variance values. 
Recently, Eilertsen \textit{et al.} \cite{eilertsen2015real} proposed an algorithm which employs histogram based locally adaptive tone curves and global tone curves to compress dynamic range and remove visible discontinuities. 
However, beyond some similarities our algorithm shares with the mentioned works, we are significantly in two ways. Firstly, all three works first split a WDR image into multiple regions and then interpolate tone curves for the pixels. The MS-Hist algorithm directly computes the per-pixel tone curve without any interpolation process. Moreover, this per-pixel tone curve computation does not increase our computation complexity because we take advantage of the integral image and integral histogram. Secondly, our final tone mapped pixel value is a weighted average of multiple tone curves of different scales where small scales are used to extract local detail and large scales are used for maintaining global brightness consistency. Works  \cite{duan2010tone} of Duan \textit{et al.} and \cite{boschetti2010high} of Boschetti \textit{et al.} both use only a single scale. Although work \cite{eilertsen2015real} of Eilerstsen \textit{et al.} adopts a 2-scale tone-curve weighting mechanism, its main purpose is to reduce the artifacts caused by the discontinuity of the local tone curves. The discontinuity artifacts in our algorithm are inherently removed by the fusion process.}

This paper is organized as follows: Section II introduces the motivation of this work. Section III explains the algorithm and optimization process in detail. Experimental results and comparisons are reported in Section IV, and we conclude the paper in Section V.

\section{Motivation}

The Human visual system (HVS) uses local adaptation to cope with large dynamic range in the real world scenarios. Local adaptation is an ability to accommodate to the level of a certain visual field around the current fixation point \cite{larson1997visibility}. Experiments carried out by Stevens and Stevens \cite{stevens1963brightness} prove that the HVS can have different responses for the same luminance level when under different backgrounds. For example, higher luminance points can be perceived darker than lower ones when they are located in different background with uniform luminance. Their results also showed that overlapped reactions in the human visual system can extend our visual response range to cope with the contrast scenes in real world. {Weber-Fechner law} \cite{hecht1924visual} states that the relation between the actual change in luminance and the perceived change is logarithmic. Logarithmic function can compress the higher luminance values as well as increase the contrast and brightness for the low luminance values.

In \autoref{Local_Adaptation}, we utilize the local adaptation and logarithimic processing features to the memorial image. The three windows' positions shown in \autoref{Local_Adaptation} (a) are randomly selected, and each window has height and width of 80 pixels. we use three different functions (shown in (b), (c) and (d)) to adapt the local dynamic ranges of the three windows. The functions map the minimum pixel value $I_{min}$ and maximum pixel value $I_{max}$ of the windows to 0 and 255, respectively. All three functions have the following form: 
\begin{equation}    
D(I) = a \cdot log(I) - b        
\label{linear log tone}
\end{equation}
where $I$ denotes the pixel value in each window. \autoref{Local_Adaptation} (d) shows the three functions in logarithmic domain. It is obvious that to adapt the three local windows the slope ($a$ value) of the three functions varies greatly. Despite of the simple compression functions used for local adaptation, the details of each window is clearly visible (also shown in \autoref{Local_Adaptation} (b), (c) and (d)). The three windows have dynamic range of 25 dB, 43 dB and 90 dB, respectively. They are much smaller than the dynamic range of the WDR image which is equal to 110.8 dB. To get a general idea about the local dynamic range level, we carried out a statistical measurement experiment. 
{We have collected over 200 WDR images from various sources including the accompany
disk of \cite{reinhard2010high}, public web page of \cite{fattal2002gradient},
\cite{debevec1997recovering}, ETHyma database\footnote{http://ivc.univ-nantes.fr/en/databases/ETHyma/}, Ward database\footnote{http://www.anyhere.com/gward/hdrenc/pages/originals.html}
and Funt database\footnote{http://www.cs.sfu.ca/~colour/data/funt\_hdr/}. We used these WDR images because they are commonly used materials for WDR and tone mapping related research.}
For each WDR image, we randomly select 200 blocks with size of 80 $\times$ 80, 120 $\times$ 120, 160 $\times$ 160, 200 $\times$ 200 and 240 $\times$ 240, respectively. We measure the average local dynamic range and standard deviation using a error bar,
the results are shown in \autoref{Fig_2}. 
Local dynamic range increases with the size of the block. However, even when the block size reaches 240 $\times$ 240, the average local dynamic range does not exceed 40 dB. 

Motivated by the local adaptation mechanism, and the measured local dynamic range feature of WDR images, we believe that even a simple tone mapping algorithm that applied to local areas could provide satisfactory tone mapping results. An apparent advantage of this local processing would be that the local area can have the full display dynamic range and therefore better preserve details. In the following, we give the details about the proposed algorithm and its optimization process.

\begin{figure}[tb]
\begin{center}
   \includegraphics[scale = 0.45]{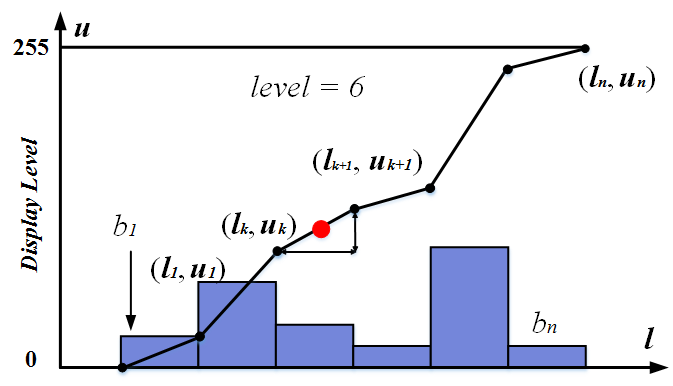}
\end{center}
   \caption{Tone mapping function based on histogram in logarithmic domain.}
\label{Hist_Equ}
\end{figure}
\section{The Proposed Algorithm}


\subsection{Local Adaptation with Histogram}
Histogram is a useful tool that takes pixel distribution into account and has been used in previously reported tone mapping algorithms such as \cite{larson1997visibility}, \cite{duan2010tone} and \cite{eilertsen2015real}. 
A histogram based tone mapping is shown in \autoref{Hist_Equ} where $l$ represents the logarithmic luminance of the pixels, and $u$ is the luminance of the displayed value. The number of bins in the histogram is a user-defined parameter $n$. If $l_{max}$ and $l_{min}$ are the maximum and minimum value, then pixels that are within the same ($l_{max}$-$l_{min})/n$ interval will fall into the same bin. The population of each bin determines the importance of the corresponding luminance levels. If one bin contains more pixels, we should assign more display levels to this bin so that details and contrast can be better preserved. Assuming $p_k$ is the population of the $k$-th bin, we assign the $k$-th bin the number of display levels that are proportional to its population and keep the display levels monotonically increasing, the maximum display level for the $k$-th bin can be calculated:
\begin{equation}  
u_k = \frac{\sum_{i=1}^{k}p_i}{\sum_{i=1}^{n}p_i } \cdot 255
\end{equation}
The cumulative sum of the histogram is used as a piece-wise linear function as shown in \autoref{Hist_Equ}. This function is locally adaptive and gives logarithmic response.

\begin{figure}[tb]
\begin{center}
   \includegraphics[scale = 0.27]{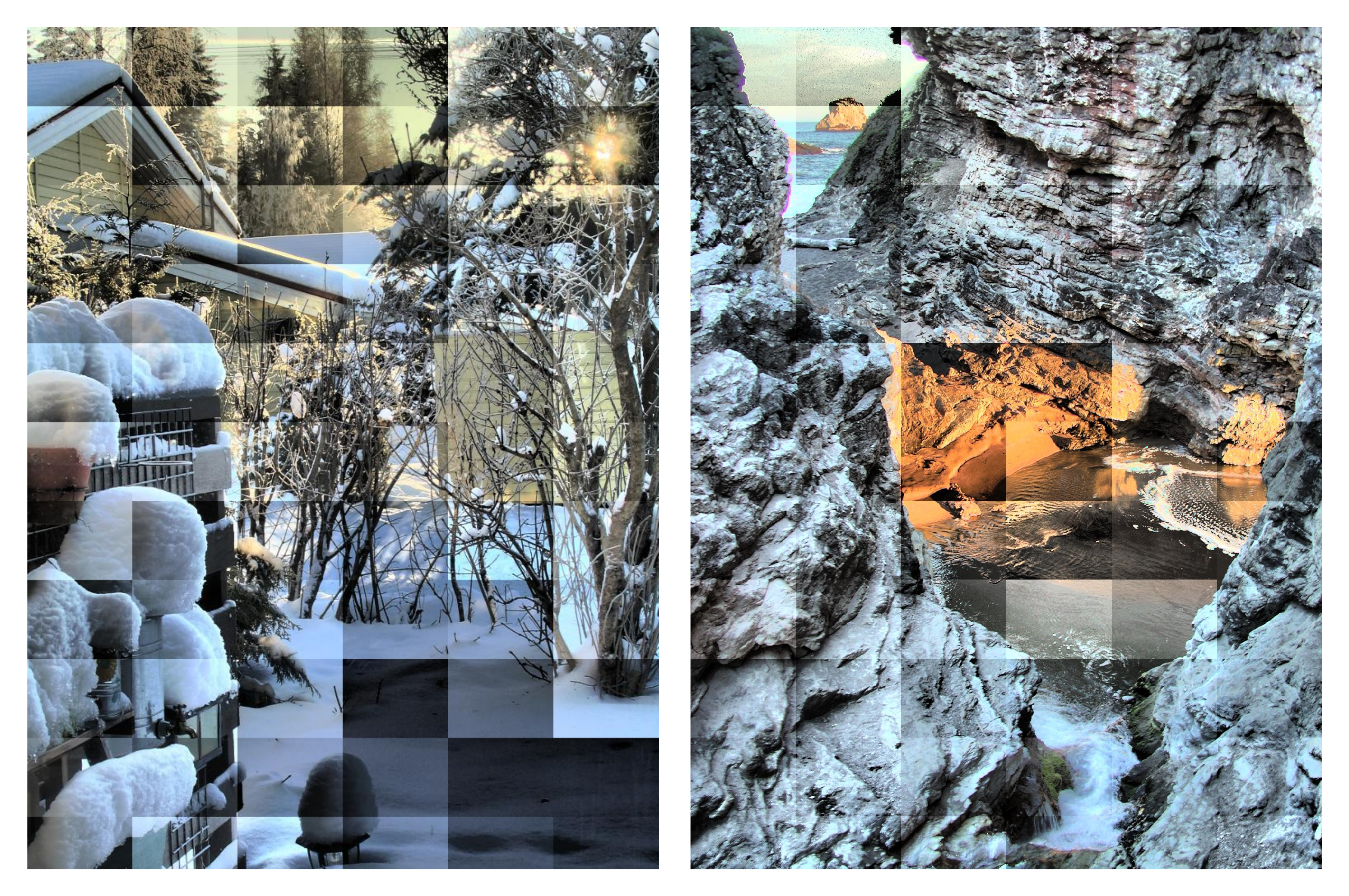}
\end{center}
   \caption{Tone mapping results using block by block method. Each
block shows good contrast and brightness. Boundary artifacts are visible at
the block edges; brightness inconsistency is also visible at some parts of the images. Radiance maps courtesy of corresponding author(s).}
\label{Fig_4}
\end{figure}

To process a WDR image locally and adaptively, the simple approach would be to divide the image into non-overlapping rectangular blocks, and compute the histogram, function and tone mapped value of each rectangular. \autoref{Fig_4} shows the tone mapping results using this method. As we expected, these images show more details and local contrast in either dark or bright regions. However, there are obvious boundary artifacts and brightness inconsistency between blocks which make the image unacceptable. In \autoref{Fig_4}, the boundary artifacts are shown as visible edges that are between any two adjacent blocks, and brightness inconsistency is shown in the snow area on the ground where pixels are mapped to dark gray rather than white. The great variation of histograms and the tone mapping functions of two adjacent blocks lead to different tone mapping functions, and this explains why pixels have different values on both sides of the boundary. The brightness inconsistency is created because local minimal value is always mapped to 0 even if the local minimal value represents high luminance. In other words, the local processing lost the global sense of the scene luminance. Besides, in uniform areas, pixels are more likely falling into the same bin of the histogram, and consequently, $a_k$ in Eq. 1 becomes a very large number, which causes small fluctuations in uniform area to be greatly exaggerated.

\begin{figure}[tb]
\begin{center}
   \includegraphics[scale = 0.39]{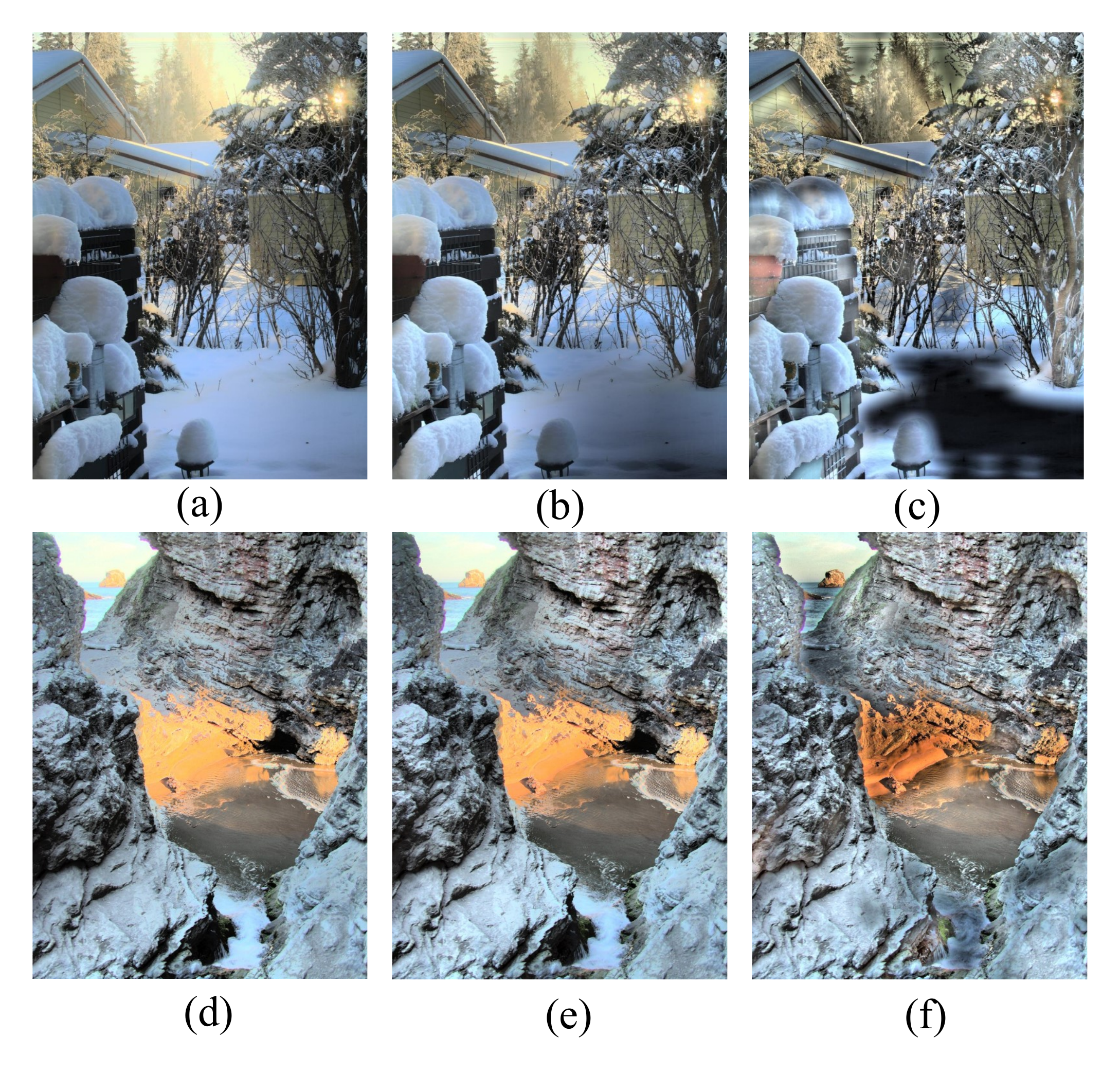}
\end{center}
   \caption{Tone mapping results using pixel by pixel method. From leftmost column to right, the block size is the size of the image size, 1/2 of the image size, 1/4 of the image size, respectively.}
\label{Fig_5}
\end{figure}

In order to remove the boundary artifacts, 
{we plan for the adjacent blocks to have as much overlap as possible so that the variation of histograms and the tone mapping functions are minimal. Hence, we adopt a pixel-by-pixel tone mapping approach in our algorithm.}
For any pixel in a WDR image, we can always find a window $w$ whose center is that pixel. The pixel distribution, tone mapping function of the window and the tone mapped value of the center pixel can be computed. If we calculate each pixel using this method, the tone mapping functions of adjacent pixels will almost be the same because the pixel distributions are highly similar between two windows which have one pixel offset. However, the local histogram and the tone mapping function will change with the scale of the window size, and therefore the tone mapped value for the centre pixel will change accordingly. \autoref{Fig_5} shows the tone mapped results using the mentioned pixel-by-pixel method of different window sizes. In \autoref{Fig_5} (a) and (d) the block size is the size of image, and the block size is 1/2 and 1/4 of the image size for \autoref{Fig_5} (b, e) and \autoref{Fig_5} (c, f), respectively. We can see that in all images, the boundary artifacts between blocks are gone. However, the brightness inconsistency issue remains for images tone mapped with scales that are smaller than the image size. In Fig. 5(b) the ground area pixels are mapped to low luminance values, and in Fig. 5(c) this area becomes totally dark.
In Fig. 5(a), the snow ground area is mapped to correct values because the window size is the same with the image size (in this case the tone mapping function is a global operator). Despite the brightness inconsistency in Fig. 5 (b) and (c), contrast and brightness of details and texture areas increase. It is obvious that smaller window scale reveal more detail of the WDR image and make certain areas of the image brighter, meanwhile the large scales can maintain more global brightness consistency.

\begin{figure}[tb]
\begin{center}
   \includegraphics[scale = 0.399]{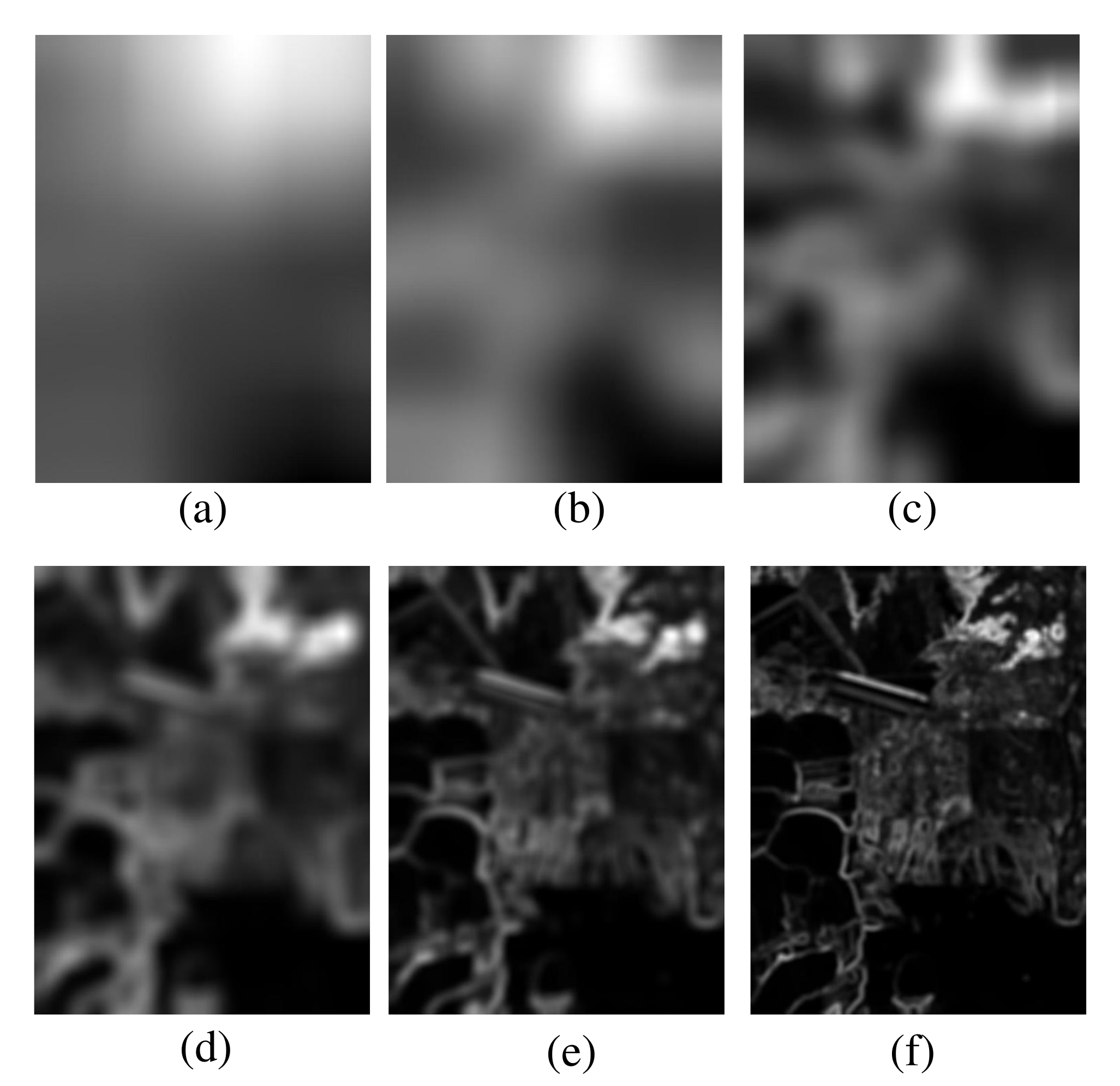}
\end{center}
   \caption{$a$ value computed in different scales. (a) $w$ is size of image. (b) $w$ is half size of image. (c) $w$ is quarter size of image. (d) $w$ is 1/8 size of image. (e) $w$ is 1/16 size of image. (f) $w$ is 1/32 size of image.
{The image is padded to handle boundary pixels.}
}
\label{Fig_6}
\end{figure}


\subsection{Multi-scale Fusion}
It is important to maintain the brightness consistency and also preserve the details during tone mapping, so that a good image can be obtained. Based on analysis of Section III. A, we propose to tone map pixels in detail and texture areas with smaller scales, while larger scales will be used for tone mapping pixels in uniform areas. In order to achieve this goal, we need to detect uniform and texture areas in WDR images first. There are a number of statistics such as entropy or measures of dispersion that can be used for detecting uniform areas. In our case, we use the following function from the recently proposed guided image filter \cite{he2010guided} to perform the task. 
%
%
\begin{equation}
a_k = \frac{\sigma_w^2}{\sigma_w^2 + \epsilon} 
\label{m-e representation}
\end{equation}
where $\sigma_w^2$ is the variance that is computed for window $w$. For window $w$, $a_k$ gives a score to its center pixel and measures if $w$ is of ``texture" or ``uniform". If the variance value $\sigma_w^2$ is much larger than $\epsilon$, the $a_k$ value will be close to 1, which indicates that the centre pixel of window $w$ is in a texture area (called high variance patch in \cite{he2010guided}) in which pixel values vary significantly. If the variance is much less than $\epsilon$, the $a_k$ value will be close to 0, and the centre pixel is regarded as belonging to a uniform area (called flat patch in \cite{he2010guided}) in which pixel values are mostly the same. Therefore, we call this factor textural score. 
The variance value and $a_k$ value change with the size of window $w$, which gives us the ability to detect uniform and texture areas under different scales. \autoref{Fig_6} shows the computed $a_k$ values of different window sizes. In these images, the brighter the pixels, the higher the corresponding $a_k$ values. It can be seen that the window size affects the values greatly. When using large window sizes, images are blurry and only skeleton can be seen. When the window size reduces, fine detail and texture gradually appear. For example, in \autoref{Fig_6} (a-c) only rough skeleton are visible, but in \autoref{Fig_6} (d-f) where the window sizes are much smaller, the leaf texture of the tree and object edges are detected. 
{Unlike other weight functions such as entropy or the one proposed in Duan \textit{et al.} \cite{duan2010tone}, the proposed weight function has a unique feature that its computation becomes very simple when taking advantage of the integral image technique. We will give more detail in the following section.}

If a pixel is in a texture area, we ideally want it to be tone mapped with a smaller window size, because it reveals more detail and contrast. If a pixel is in a uniform area, we ideally want it to be tone mapped with a large window size, because it can mantain brightness consistency and brings no artifacts. We fullfill this goal with the following equation:
\begin{equation}
u = \frac{\sum_{i=1}^{s-1}a_{w_i}^iu_{w_i}}{\sum_{i=1}^{s-1}a_{w_i}^i} 
\label{histogram-equal-adjusted}
\end{equation}
For each pixel in WDR image, $u_{w_i}$ and $a_{w_i}$ represent the tone mapped value and the texture score that are computed under scale $w_i$. $s$ is the number of scales that are used.
Considering a pixel in a uniform area, $a_{w_i}$ will be close to $0$, and $a_{w_i}^i$ will approach $0$ more rapidly because of the exponent. This could greatly reduce the weight of the corresponding $u_{w_i}$ value. However, if a pixel is in a uniform area, then $a_{w_i}$ will be close to 1, $a_{w_i}^i$ will also be close to 1, which gives more weight to detail from small scales.
{The fusion function intuitively assigns more weight to large scales and less weight to small scales using the different exponent values of Eq. 6. Tone curves generated in larger scales are more global, they don't introduce artifacts as we have shown in Fig. 5. Small scales are more local, and the corresponding images are more likely to have artifacts. The biased weights of large and small scales minimizes any visible artifacts. }

\subsection{Optimization}

The proposed algorithm needs to compute variance and histogram of multiple scales at each pixel location. It will be a time-comsuming process, especially under circumstances that the resolutions of current WDR images are mostly over mega-pixels. Optimizing the process is inevitable if an acceptable processing time is required. Here, we adopt integral image \cite{viola2004robust} and integral histogram \cite{porikli2005integral} to reduce complexity and processing time of the variance and histogram computation, respectively.

The variance value $\sigma_w$ can be computed by:
\begin{equation}
\sigma_w = \frac{\sum_w{i^2} - (\sum_w{i})^2}{|w|^2} 
\label{m-e representation}
\end{equation}
where $i$ is the pixel of the WDR image, $|w|^2$ is the number of pixels in window $w$. We accelerate the above calculation with the help of integral image. If $I$ is the integral image of the WDR image, then the value at any location $I(x, y)$ is calculated by:
\begin{equation}
{T}(x, y) = \sum_{x' \leq x, \ y' \leq y}i(x', y') 
\end{equation}
A great feature of integral image 
{$T$} 
is that summation of any rectangular region in the original image can be computed efficiently in a single pass. For example, if there are four points $A (x_0, y_0)$, $B (x_1, y_0)$, $C (x_1, y_1)$ and $D (x_0, y_1)$ in image $i$, the rectangular that is enclosed by the four points is equal to:
\begin{equation}
 \sum_{\substack{x_0 < x_1 \le x_1 \\ y_0 < y <y_1}}i(x, y) = {T}(A) + {T}(C) - {T}(B) - {T}(D) 
\label{RGBtoGray}
\end{equation}
For fast computation of Eq. 9, we can first get the integral image of $i$ and $i^2$. Then, the summation of $\sum i^2$ and  $\sum i$  of any window $w$ can be replaced with simple addition and subtraction operations as shown in Eq. 11. 

To facilitate the computation of histogram of any window size, 
\begin{figure}[tbh]
\begin{center}
   \includegraphics[scale = 0.35]{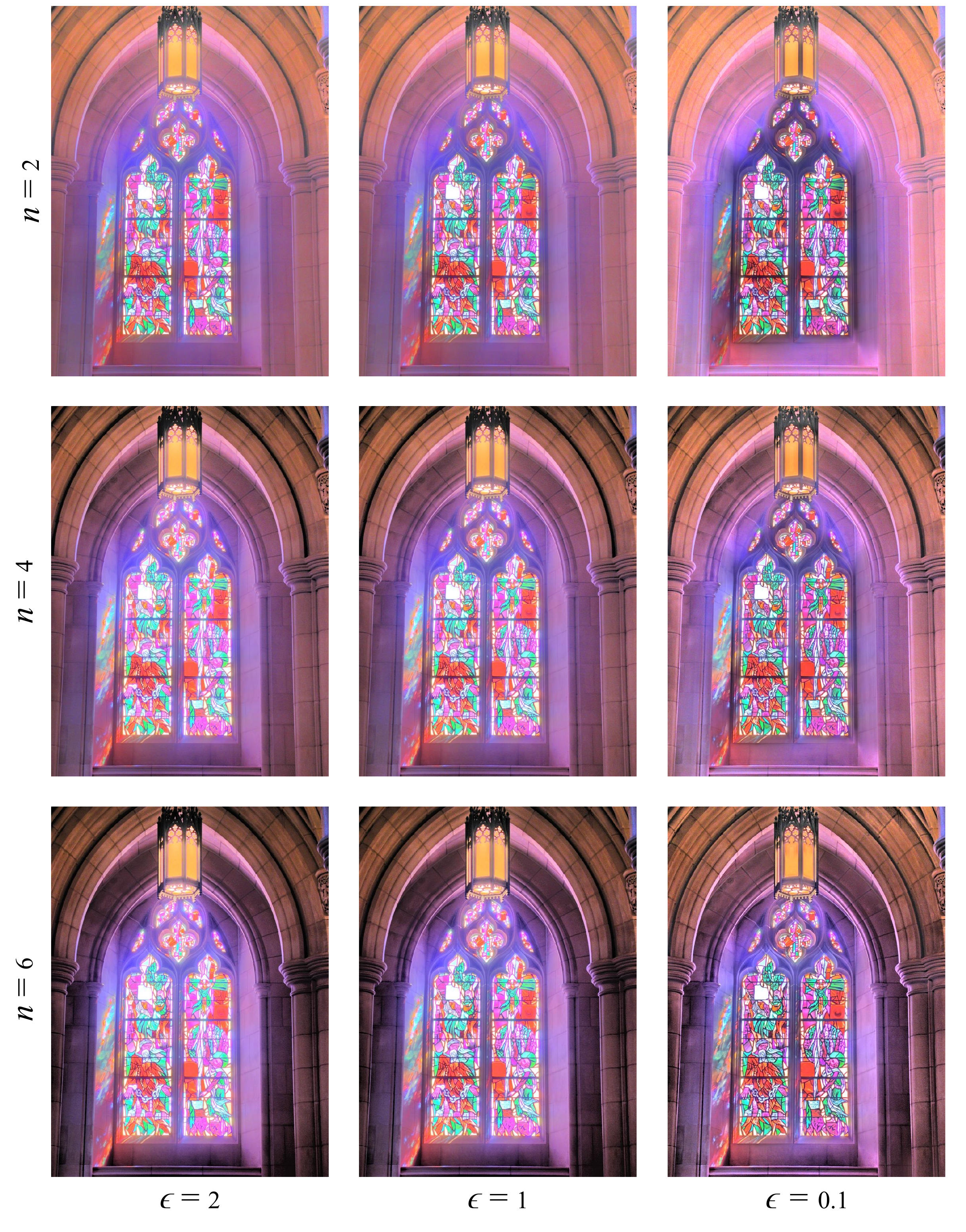}
\end{center}
\caption{Tone mapped images obtained with various $n$ and $\epsilon$ values.}
\label{ParamsChange}
\end{figure}
we first construct an $n$-channel integral histogram $H$ where the $k$-th channel $H_k$ is an integral image. It is computed by:
\begin{equation}
H_k(x, y) = \sum_{\substack{x' \leq x, \   y'\leq y   }} i_k(x',y') 
\end{equation}
where
\begin{equation}
i_k (x', y')=\left\{
\begin{array}{rcl}
1      &      & {i(x, y)      \in      b_k}\\
0      &      & {i(x, y)      \notin b_k}
\end{array} \right.
\end{equation}
$b_k$ is the k-th bin of the histogram. After the $H$ is computed, population of the $i$-th bin can be computed instantly using also simple additions and subtractions: 
\begin{equation}
 p_i = H_i(A) + H_i(C) - H_i(B) - H_i(D) 
\label{RGBtoGray}
\end{equation}
After $p_i$ values are computed, tone mapping function can be easily computed by using Eq. 2.

Integral image and integral histogram only need to be computed once for the use of all different scales. Subsequent computations are mostly simple addition and subtraction operations {which can be carried out in a matrix parallel fashion.}
Additional parallelization can further optimize the proposed algorithm because the computation of different scales are independent and can also be computed in parallel. {The use of integral image and integral histogram greatly reduces the per-pixel tone curve computation burden, because the majority of the algorithm is computed in in parallel.}


\section{Implementation and Experimental Results}
\subsection{Implementation}
The WDR color images have red, green, blue three channels, but our algorithm operates on the luminance channel. Consequently, we first convert the HDR image to luminance image by using \autoref{RGBtoGray} below:
\begin{equation}
 L = 0.299R + 0.587G + 0.114B 
\label{RGBtoGray}
\end{equation}
The color information is restored by using the following equation \cite{fattal2002gradient}:

\begin{equation}
C_{out} = (\frac{{C_{in}}}{L_{in}})^{sat} \cdot L_{out}
\label{Color}
\end{equation}
where $C = R, G, B$ represents the red, green and blue channel, respectively. $L_{in}$ and $L_{out}$ denote the luminance before and after the tone mapping. Parameter $sat$ controls the color saturation degree of the result image. 
{
If it is set too small, the result image looks pale, but if it is set too large, the color will become over saturated. \cite{fattal2002gradient} found that when this parameter is between 0.4 and 0.6, the results are satisfactory, Gu \textit{et al.} \cite{gu2013local} set it as 0.6 in their experiment. In our experiments, we also set $sat = 0.6$ because it gives good color performance for most images.}

The number of scales $s$ is the most important parameters in the proposed algorithm. As we have stated previously, in order to maintain the image brightness consistency, the block size of the largest scale should be the same as the image size. To obtain possible details in different scales, we adopt the popular image pyramid method. Therefore, for any two adjacent scales, the block size $w_{i+1}$ is computed as $w_i/2$.

In a lot of algorithms, $64 \times 64$ are considered as a small enough block size for image processing. Hence, we chose our number of scales so that the smallest block size is equal to or less than $64 \times 64$. Considering a $2000 \times 2000$ WDR image, 6 scales are enough to make the smallest block size satisfy our criteria.

Two other free parameters are involved in our algorithm: the number of bins $n$, and the regularization term $\epsilon$. The effect of these parameters for a tone mapped image is shown in \autoref{ParamsChange}. Nine tone mapped images are presented in a matrix with $n$ varying vertically and $\epsilon$ varying horizontally. The overall image contrast increases with the increase of $n$, while more local details are revealed with the decrease of $\epsilon$. We find values for $n = 5$ and $\epsilon = 0.1$ to usually produce satisfactory results, good brightness while preserving local contrast and details.

\subsection{Customized Application}
\begin{figure}[tbh]
\begin{center}
   \includegraphics[scale = 0.12]{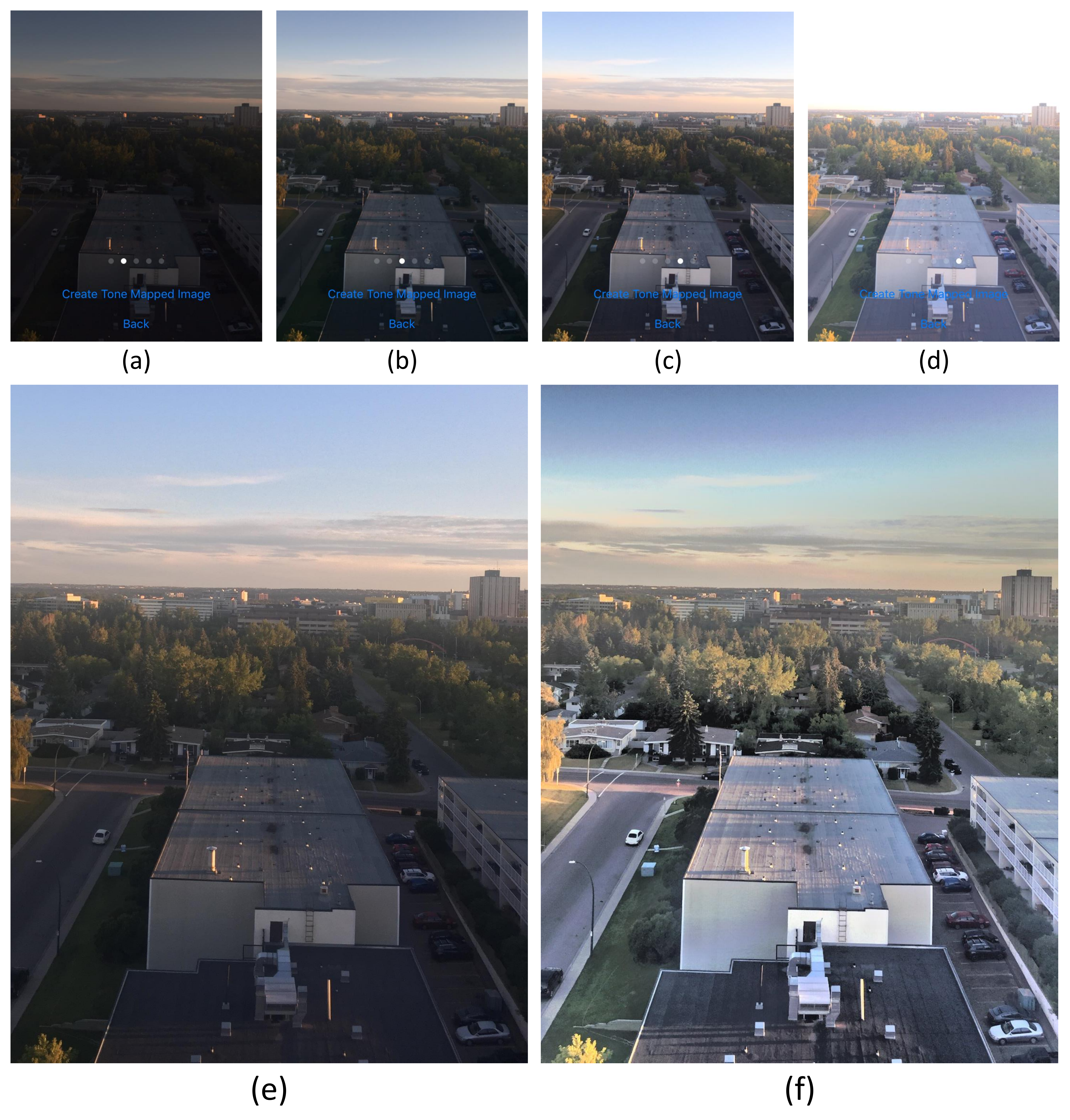}
\end{center}
   \caption{WDR image captured by phone camera and our application. (a-d) LDR image captured by phone camera. (e) merged image with iPhone 6's built-in HDR feature.
   (f) WDR image captured by our developed application.}
   \label{Fig_9}
\end{figure}
We implemented our algorithm in both Android and iOS platforms. The developed application  is called \textit{CaptureWDR}. The application first captures four sequential images under different exposures and merges them into a WDR image with Paul Debevec's method \cite{debevec2008recovering}. The WDR image is then tone mapped with the proposed algorithm. \autoref{Fig_9} (a)-(d) show four LDR images taken under different exposures with a phone camera. The exposure time increase steadily that the high lighted cloudy sky and the low luminance roof are both captured in different images. \autoref{Fig_9} (e) shows the  image captured with iPhone 6's built-in HDR feature. \autoref{Fig_9} (f) shows the tone mapped image of our application.
Our result is significantly brighter than \autoref{Fig_9} (e) and shows more detail, especially the roof and the parking lot on the right side of the image. In the iOS and Android implementation, we set the number of scales $s = 5$, number of bins $n = 5$ and $\epsilon = 0.1$ as default values. The response time for our application is less than a second on a iPhone 6 including WDR merging, tone mapping and displaying. We haven't adopted any parallelism or GPU computing in our program yet, the response time will reduce significantly after any parallel programming. We have tested our application in various lighting situations such as sunny outdoor, dark indoor and shaded areas. The resulting images are satisfactory and exhibit good contrast and brightness.


\begin{figure*}[tbh]
\begin{center}
   \includegraphics[scale = 0.38]{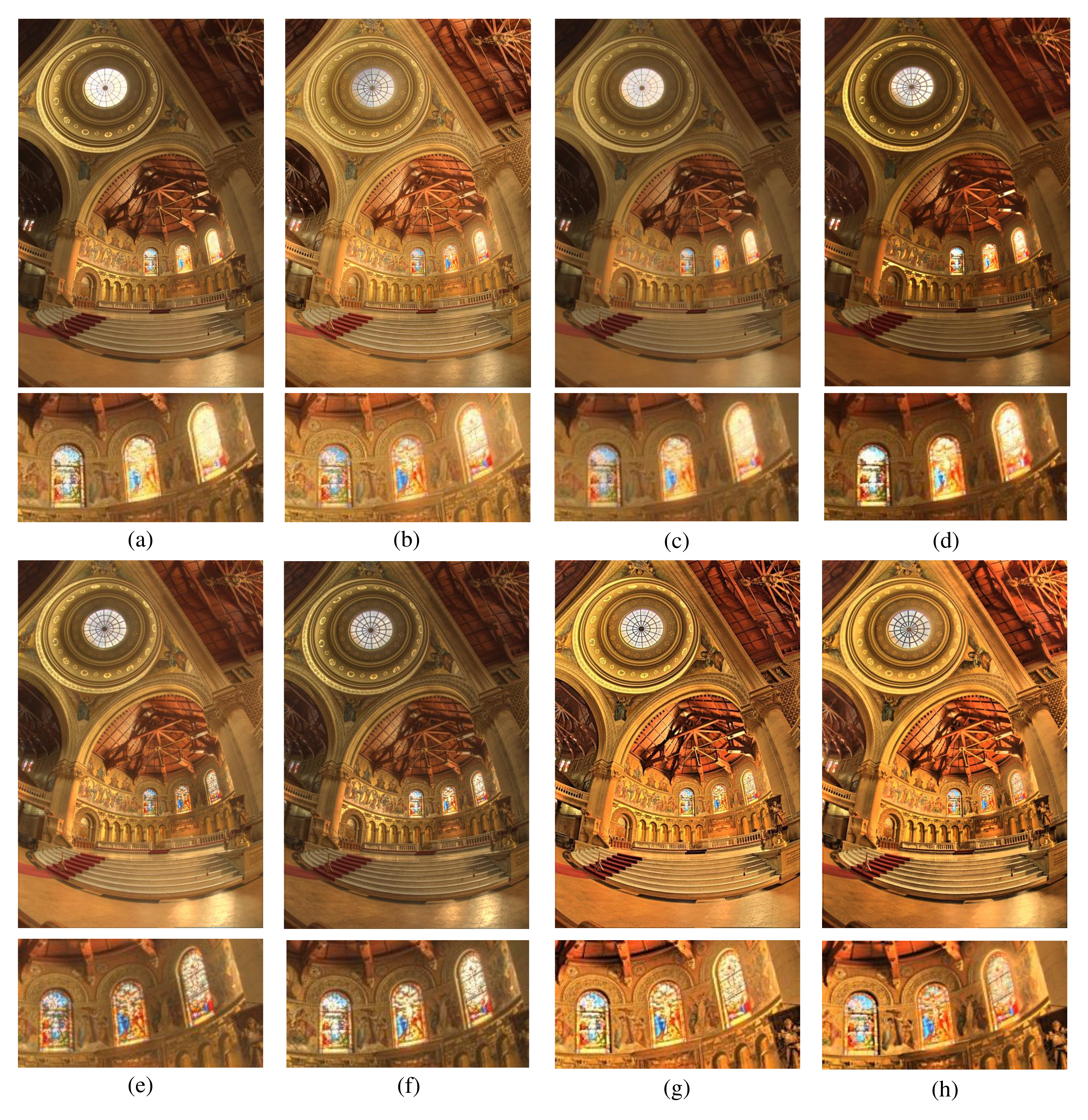}
\end{center}
   \caption{Comparison of the reproduced memorial church WDR image between our algorithm and other algorithms. (a) Result taken from \cite{reinhard2002photographic}. (b) Result taken from \cite{fattal2002gradient}. (c) Result taken from \cite{drago2003adaptive}. (d) Result of \cite{durand2002fast}. (e) Result of \cite{meylan2006high}. (f) Result of \cite{paris2015local}. (g) Result of \cite{gu2013local}. (h) Result of our proposed MS-Hist algorithm.}
\label{compare-memorial}
\end{figure*}
\subsection{Comparison Results}

Our proposed algorithm is compared with other reported works. We first choose the famous memorial church image for comparison because it is one of the most commonly used image for testing tone mapping algorithms.
\autoref{compare-memorial} shows the image tone mapped with eight different tone mapping algorithms including ours. The seven other algorithms are state-of-the-art tone mapping algorithms and they all show effective tone mapping efforts. In the eight images, only (b), (g) and our result show averagely bright scenes, other images are dark, and the details of the upper left corner are lost. In images (b), (d) and (f), the upper left corner are visible, but only in images of (g) and (h) the details are completely presented. The contrast near 
the window area of the memorial church image is particularly challenging; this part is zoomed and also showed in \autoref{compare-memorial}. In images (a)-(d), this area is mostly saturated, especially on the rightmost window: the paintings between the windows suffer from loss of contrast. In images (e)-(g), the window area {is} no longer saturated, and the painting on the rightmost window can be seen clearly. Among those four images, ours shows the best contrast, especially on the paintings between the windows.
 
The memorial image example in \autoref{compare-memorial} shows that our algorithm can produce visually bright and high contrast images. Although our image may suffer some loss in naturalness, but it is clear that this is in exchange of the visibility which is more important and critical. The following assessment alone with TMQI\cite{yeganeh2013objective} score in later experiments validate that tone mapped image overall in high quality.
In the following, we use three simple quality measures to {assess} the performance of our tone mapping algorithms: the brightness value $\alpha$, the sharpness value $\beta$ and the standard deviation value which are used to measure the image brightness, details and contrast respectively.
The brightness value $\alpha$ and sharpness values $\beta$ are computed using the following equation:
\begin{equation}
{
\alpha =\frac{1}{N} \sum F, \ \ \ \beta = \sum |\nabla F|}
\label{local contrast}
\end{equation}
where $N$ is the number of pixels in final tone mapped image 
{$F$}.
The comparison results for the images in \autoref{compare-memorial} are shown in \autoref{overall compare}. The results are consistent with our visual experience. Our tone mapped image obtains the highest sharpness and contrast values. In the psychophysical experiment, both Gu \textit{et al}.'s and our algorithm achieve high scores.
\begin{table}
\centering
\vfill
\caption{\textsc{Quantitative Measurements On \autoref{compare-memorial}}}
\begin{tabular}{ |l | l | l | l| }
\hline
\hline
Image & \tabincell{l}{ Sharpness} &  Brightness &  Contrast\\
\hline
 Fig. 9(a) Reinhard \cite{reinhard2002photographic} & 8.4697& 91.3766 & 53.7200 \\
 Fig. 9(b) Fattal \cite{fattal2002gradient} & 13.0405 & \textbf{117.1412} & 58.1158 \\ 
 Fig. 9(c) Drago \cite{drago2003adaptive} &5.8122 & 85.4665& 47.7418\\  
 Fig. 9(d) Durand \cite{durand2002fast} & 7.6244 & 73.7916 & 48.5908 \\    
 Fig. 9(e) Meylan \cite{meylan2006high} & 8.461 & 104.6918 & 54.3960 \\
 Fig. 9(f) Paris \cite{paris2015local} & 9.4782  &  85.2719 & 47.7444 \\
 Fig. 9(g) Gu \cite{gu2013local}   & 13.1790    & 110.0256 & 68.5587\\ 
 \tabincell{l}{Fig. 9(h) Currently prop- \\osed algorithm MS-Hist} & \textbf{14.8941}  & 115.1672 & \textbf{70.6124}\\
  \hline
\end{tabular}
\label{overall compare}
\end{table}

Objective assessment of tone mapping algorithms is important. In recent years, indexs such as tone-mapped image quality index (TMQI) \cite{yeganeh2013objective}, feature similarity indexes
for tone-mapped images (FSITM) \cite{nafchi2015fsitm}, blind tone-mapped quality indexes (BTMQI) \cite{gu2016blind}, HDR-VDP \cite{mantiuk2011hdr}, DRIM \cite{gu2013local} were proposed to provide a single score for evaluating tone mapping algorithms. These indexes mostly consider one or two aspects such as naturalness and structureness of the tone mapped images and give corresponding scores. However, evaluation of images are also a psychophysical process determined by human observers, and it can hardly be fully described by only few metrics. {In our experiment, we use both psychophysical experiment and objective assessment to evaluate different tone mapping algorithms.} We conduct psychophysical experiment first and then make objective assessment with TMQI.


We select 20 images for a psychophysical experiment. These images are standard WDR radiance maps that are widely used in tone mapping algorithm tests. All images and corresponding WDR files can be found in our project site\footnote{https://github.com/jieyang1987/Tone-Mapping-Based-on-Multi-scale-Histogram-Synthesis}. In our experiment, {we give each participant a website address\footnote{https://surveyhero.com/c/53b8aa3} which links to our on-line survey. The participant doesn't have any prior knowledge on any of the tested algorithms. There are no constrains on where and when they should take the survey. The participants can also choose whatever device that comes in handy to do the survey such as mobile phone, tablet or computer screen. Unlike some experiments that are done in a controlled environment, our experiment gives a more robust result reflecting how the algorithm performs in real life. In the on-line survey,}
anticipates were shown 20 questions, each question is related to four images that had been tone mapped with different algorithms. The four images were randomly marked with (a), (b), (c) and (d). Observers were asked to choose the image that they prefer the most. Observers were allowed to choose multiple images if they thought the images were equally pleasant. A total number of 129 volunteers anticipated to the psychophysical experiment. The results are summarized in \autoref{psychophysical_experimentation}. Overall, we got the most votes for 14 images and a total number of 877 votes. Gu \textit{et al.}'s algorithm \cite{gu2013local} won in four images, and it got 544 votes. After the experiment, we found that observers are more likely to choose images exhibit more brightness and contrast. 
{We use TMQI index score to assess the 20 images that are used in the psychophysical experiment. The results are listed in Table III. In the 20 images, our algorithm gets highest TMQI scores for 10 images. Paris \textit{et al.}, Gu \textit{et al.} and Durand \textit{et al.} get highest TMQI scores for 7, 2 and 1 images, respectively. MS-Hist also achieves the highest average TMQI value among the four algorithms. The TMQI score gives rough idea of the overall quality, but it is not always precise. For example, \textit{AtriumMorning} image tone mapped by our algorithm is the most preferable result, but it has the lowest TMQI value. We show all the tested image in our supplementary material.}  

\begin{table}
\centering
\vfill
\caption{\textsc{Statistic Results of Psychophysical Experimentation}}
\begin{tabular}{ l  c  c  c  c}
\hline
\hline
Image & \tabincell{c}{Durand \textit{et al.} \\ \cite{durand2002fast}} & \tabincell{c}{Paris \textit{et al.} \\ \cite{paris2015local}} & \tabincell{l}{Gu \textit{et al.} \\ \cite{gu2013local}} & Proposed \\
\hline
 AtriumMorning         & 2 & 8 & 30 & \textbf{84}   \\ 
 AtriumNight           & 4 & 16 & 40 & \textbf{56}  \\    
 belgium			   & 1 & 3 & \textbf{56} & 43  \\
 cathedral             & 13& 30 & \textbf{34}  & 28 \\
 crowfoot			   & 6 & 33 & 13 & \textbf{42}  \\
 designCenter          & 4 & 6  & \textbf{43} & 42  \\
 groveD                & 8 & 5  & 31  & \textbf{47} \\
 memorial              & 17& 20 & 29  & \textbf{31} \\
 moraine1              & 2 &  29 & 14 & \textbf{45} \\
 moraine2              & 1 & 10  & 36 & \textbf{46} \\
 orion 				   & 9 & 17  & 29 & \textbf{38} \\
 tmN	               & 19& 13  & \textbf{33} & 29 \\
 vernicular            & 5 & 10  & 23 &\textbf{54}  \\
 vinesunset            & 2 & 21  & 31 &\textbf{41}  \\
 rend01                & \textbf{37} & 22 & 6 & 24 \\
 Rockies3b             & 3 & 16  & 12 & \textbf{56} \\
 moto                  & 19 & \textbf{42} & 2 & 27 \\
 tinterna              & 2 & 16 & 12 & \textbf{61} \\ 
 nancy\_cathedral      & 3 & 6 & 38 & \textbf{44} \\
 garage                & 5 & 15  & 32 & \textbf{39} \\ \hline \hline 
 Total                 &  162 & 338 & 544 & \textbf{877}\\ 
 \hline 
\end{tabular}
\label{psychophysical_experimentation}
\end{table}

\begin{table}

\centering
\vfill
\caption{\textsc{TMQI Scores for the Tested Images}}
\begin{tabular}{ l  c  c  c  c}
\hline
\hline
Image & \tabincell{c}{Durand \textit{et al.} \\ \cite{durand2002fast}} & \tabincell{c}{Paris \textit{et al.} \\ \cite{paris2015local}} & \tabincell{l}{Gu \textit{et al.} \\ \cite{gu2013local}} & Proposed \\
\hline
 AtriumMorning         & \textbf{0.9722} & 0.9315 & 0.8797 & 0.8669   \\ 
 AtriumNight           & 0.8316 &\textbf{0.9306} &   0.9194&   0.9146\\    
 belgium			   & 0.8183 & 0.8650 &  0.9400& \textbf{0.9428}  \\
 cathedral             & 0.8586	& 0.8933 & 0.8900 & \textbf{0.9154} \\
 crowfoot			   & 0.8187 & \textbf{0.9119}  &0.8705  & 0.8897  \\
 designCenter          & 0.7334 & 0.7984 &  0.8753 & \textbf{0.9529}  \\
 groveD                & 0.9236 & 0.9160   & 0.8350  & \textbf{0.9601} \\
 memorial              & 0.8689	& \textbf{0.9034} &  0.8521  & 0.8476 \\
 moraine1              & 0.8317 &  0.9025  & 0.8622 & \textbf{0.9027} \\
 moraine2              & 0.8076 &  0.8666  & \textbf{0.9373}& 0.9230 \\
 orion 				   & 0.7016 &  0.7014 & 0.7655 & \textbf{0.7817} \\
 tmN	               & 0.7898	&  0.8435 &  0.8054 & \textbf{0.8962} \\
 vernicular            & 0.8681 &  \textbf{0.9440} &  0.9170 &0.9306  \\
 vinesunset            & 0.8130 &   0.8628 &  0.7847 &\textbf{0.8821}  \\
 rend01                & 0.9366 & \textbf{0.9400} & 0.7545  & 0.9074 \\
 Rockies3b             & 0.8328 &  0.8957 &  0.8422 & \textbf{0.9385} \\
 moto                  & 0.8140 & \textbf{0.8692} &  0.8200 & 0.7619 \\
 tinterna              & 0.8586 & \textbf{0.9589}&   0.9431 & 0.9566\\ 
 nancy\_cathedral      & 0.7666 &  0.8365 & 0.9612 & \textbf{0.9657} \\
 garage                & 0.7551 &  0.8379 &  \textbf{0.9656} &0.9605 \\ \hline \hline 
 Average                 & 0.8300 & 0.8805 & 0.8710 & \textbf{0.9048}\\ 
 \hline  
\end{tabular}
\label{objective_score}
\end{table}



\section{Conclusion and Further Work}
In this paper, we have presented a novel tone mapping algorithm based on multiple-scale histograms. Histograms of different scales correspond to different tone mapping functions where functions of large scale histograms are used to maintain image brightness consistency and functions of small scales are used to preserve local details. A fusion function was proposed in order to take advantage of the large scale functions and small scale functions so that both WDR image brightness consistency and local details are both kept in the final tone mapped image. The proposed algorithm is also optimized to reduce computation complexity and processing time. We have compared our algorithm with some other tone mapping algorithms through objective and psychophysical experiments. The results have shown that our algorithm can produce appealing images with high brightness and high contrast. 

Our further work would be to improve the user interface, image stabilization, implement GPU version of the algorithm and apply the algorithm to fields such as bio-medical imaging, infrared imaging and security applications.


\section{Acknowledgement}
The authors would like to thank the Alberta
Innovates Technology Futures (AITF) and Natural Sciences and
Engineering Research Council of Canada (NSERC) for supporting this research. The authors would like to thank Nasir Mohamed Osman and Douglas Michael McDonald for developing the Android and iOS applications. The authors also would like to thank Dr. Alain Hor\'e for providing valuable suggestions on writing of this paper. At last, the authors would like to thank all participants of the psychophysical experiment.

{\small
\bibliographystyle{ieeetr}
\bibliography{egbib}
}

\end{document}